\documentclass[runningheads]{llncs}
\usepackage[T1]{fontenc}
\usepackage{graphicx}
\usepackage{amsfonts}
\usepackage{amsmath}
\usepackage{dsfont}
\usepackage{booktabs}
\usepackage{graphicx}
\usepackage{xcolor}
\usepackage{arydshln}
\usepackage{multirow}
\begin{document}
\title{Event-Priori-Based Vision-Language Model for Efficient Visual Understanding}
\titlerunning{Event-Priori-Based Vision-Language Model}
%
\author{
Haotong~Qin\inst{1} \and
Cheng~Hu\inst{2}\thanks{Corresponding Author} \and
Michele~Magno\inst{1}
}
\authorrunning{
H. Qin et al.
}
%
\institute{
Center for Project-Based Learning D-ITET, ETH Zürich, Switzerland\\
\email{\{haotong.qin,michele.magno\}@pbl.ee.ethz.ch} \and
State Key Laboratory of Industrial Control Technology, Zhejiang University, China\\
\email{22032081@zju.edu.cn}
}
\maketitle              
\begin{abstract}
Large Language Model (LLM)-based Vision-Language Models (VLMs) have substantially extended the boundaries of visual understanding capabilities. However, their high computational demands hinder deployment on resource-constrained edge devices. A key source of inefficiency stems from the VLM's need to process dense and redundant visual information. Visual inputs contain significant regions irrelevant to text semantics, rendering the associated computations ineffective for inference. This paper introduces a novel Event-Priori-Based Vision-Language Model, termed \textbf{EP-VLM}. Its core contribution is a novel mechanism leveraging motion priors derived from dynamic event vision to enhance VLM efficiency. Inspired by human visual cognition, EP-VLM first employs event data to guide the patch-wise sparsification of RGB visual inputs, progressively concentrating VLM computation on salient regions of the visual input. Subsequently, we construct a position-preserving tokenization strategy for the visual encoder within the VLM architecture. This strategy processes the event-guided, unstructured, sparse visual input while accurately preserving positional understanding within the visual input. Experimental results demonstrate that EP-VLM achieves significant efficiency improvements while maintaining nearly lossless accuracy compared to baseline models from the Qwen2-VL series. For instance, against the original Qwen2-VL-2B, EP-VLM achieves 50\% FLOPs savings while retaining 98\% of the original accuracy on the RealWorldQA dataset. This work demonstrates the potential of event-based vision priors for improving VLM inference efficiency, paving the way for creating more efficient and deployable VLMs for sustainable visual understanding at the edge.
\keywords{Vision-language model  \and Event data \and Visual understanding}
\end{abstract}
\section{Introduction}
Vision-Language Models (VLMs)~\cite{bai2025qwen2,wang2024qwen2,liu2023visual,radford2021learning,guo2025seed1} have revolutionized visual understanding by unifying visual and linguistic modalities. These models often build upon the strong reasoning and language understanding capabilities of Large Language Models (LLMs)~\cite{yang2025qwen3,meta2025llama,guo2025deepseek,hurst2024gpt,liu2024deepseek}, extending them to interpret and process visual information through large-scale pre-training. VLMs, powered by sophisticated LLM backbones, demonstrate remarkable capabilities across a wide array of applications, including image captioning~\cite{bai2025qwen2}, visual question answering~\cite{antol2015vqa}, object detection~\cite{kuo2022f}, and multimodal content generation~\cite{he2024llms}. They are finding use in domains such as autonomous driving~\cite{baumann2025enhancing}, robotics~\cite{song2023llm}, healthcare imaging~\cite{li2024integrated}, and interactive AI assistants~\cite{guan2023intelligent}. This success has also spurred the exploration of their deployment on edge, such as on-device visual search, real-time surveillance analysis, and human-robot interaction on portable devices.

\begin{figure}[t]
\centering
\includegraphics[width=0.93\textwidth]{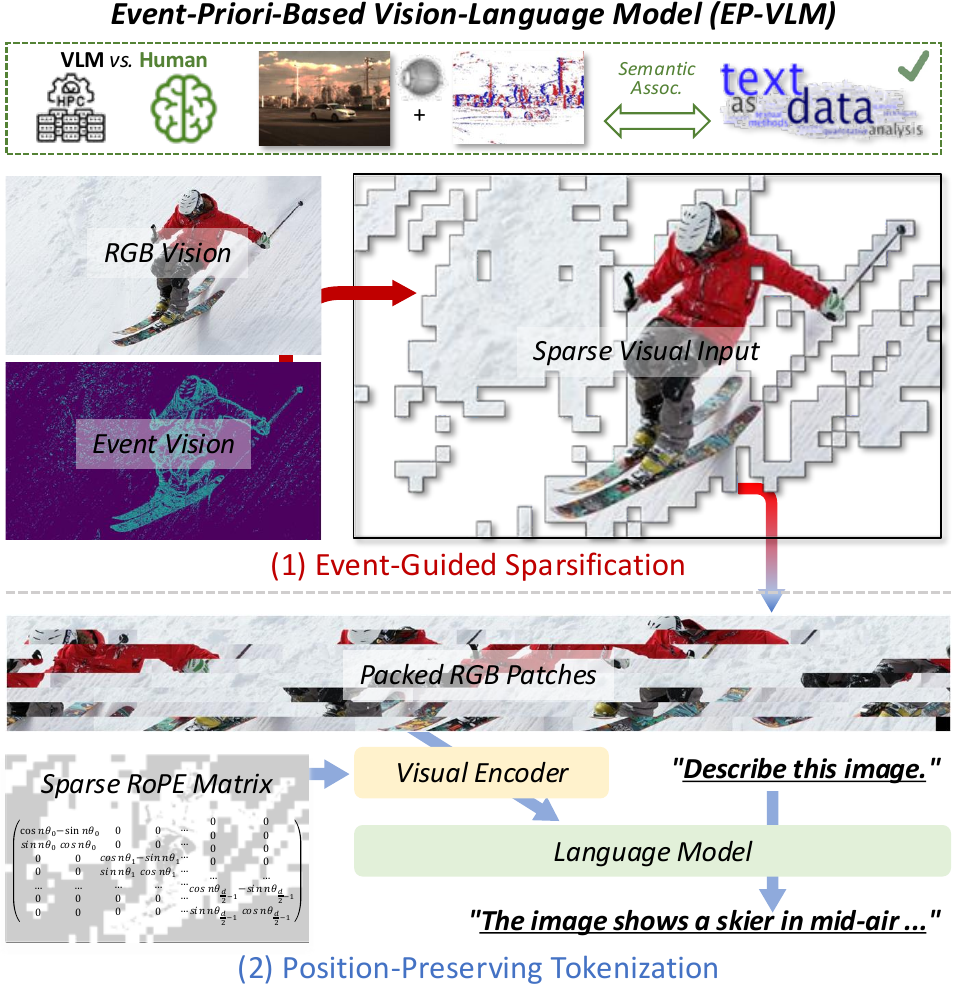}
\caption{Overview of EP-VLM.}
\label{fig:overview}
\end{figure}

However, deploying VLMs, which inherently carry the computational load of integrated LLMs, in resource-constrained edge environments presents significant challenges. These include high latency, limited deployment feasibility, and excessive energy consumption. Existing mainstream VLM frameworks, such as Qwen-VL~\cite{bai2025qwen2} and LLaVA~\cite{liu2023visual}, exhibit substantial resource demands. This is partly attributable to the large parameter counts in their underlying LLMs. For instance, large-scale VLMs can have tens of billions of parameters, while smaller variants like LLaVA-1.5-7B still rely on multi-billion parameter LLMs. These models typically require significant GPU memory and computational power, often exceeding 10GB for inference and hundreds of GFLOPs per token generated. In contrast, mainstream mobile devices typically feature limited RAM, significantly lower computational power, and strict battery constraints. Although compression methods targeting both visual frontends and LLM backends, such as pruning~\cite{ma2023llm,sun2023simple}, quantization~\cite{qin2024accurate,huang2024billm,huang2024empirical,chen2024db}, and knowledge distillation~\cite{xu2024survey}, have been explored, they often compromise accuracy on complex visual tasks. These approaches generally fail to achieve Pareto optimality between accuracy and efficiency. Consequently, there is an urgent need for computationally efficient VLM mechanisms to enable sustainable visual understanding at the edge.

Existing Vision-Language Models (VLMs) achieve visual-language understanding by processing directly encoded input information, namely RGB images (or video frames) and text, during inference. However, for the former, due to visual information being sparsely distributed within dense pixel arrays, a significant amount of semantically irrelevant background information is continuously introduced into the VLM's computational inference. Processing of visual inputs occurs within both the visual encoder and the LLM backbone: the former encodes the input into tokens, while the latter computes and manages both visual and text tokens. Consequently, vision-related computation constitutes a substantial portion of the VLM's overall computational load. Depending on the architecture and input resolution, computation directly attributable to visual input can even account for over 90\% of the total computational cost during image caption generation. Therefore, the efficiency wasted on semantically irrelevant visual information remains a critical, yet often overlooked, bottleneck in VLM-based visual content understanding.

To address this challenge, this paper proposes an \textbf{E}vent-\textbf{P}riori-Based \textbf{V}ision-\textbf{L}anguage \textbf{M}odel, namely \textbf{EP-VLM}, for efficient visual understanding (Fig.~\ref{fig:overview}). Unlike existing VLMs, which rely on expensive internal computation for cross-modal understanding, humans leverage motion information from biological visual cues as prior knowledge to assist the brain in achieving efficient visual cognition~\cite{berry1999anticipation}. Inspired by this biological mechanism, we propose a novel approach utilizing motion priors derived from dynamic event vision to sparsify visual information, thereby enhancing the efficiency of visual understanding while maintaining accuracy. Firstly, we introduce event vision data temporally aligned with RGB image sampling and employ it as prior knowledge to guide patch-wise sparsification of the RGB visual input. This means we only process visual regions indicated by the event data as salient or changing. Subsequently, we adapt the visual encoder within the VLM architecture to handle this event-guided unstructured sparse visual input, ensuring that positional encoding is correctly propagated and matched to enable accurate comprehension of the sparse visual information while saving computation.

Our preliminary experiments demonstrate that the proposed EP-VLM achieves significant efficiency improvements compared to its baseline, Qwen2-VL. For instance, EP-VLM maintains 98\% of the baseline's response accuracy on the RealWorldQA dataset~\cite{XAI2024Grok2} while achieving a reduction of approximately 50\% in computational FLOPs and MACs. Concurrently, our qualitative study indicates that the event vision priors in EP-VLM effectively preserve the semantic content and spatial relationships of the visual input. Thus, EP-VLM leverages event vision to significantly enhance VLM efficiency, paving the way for deploying more capable and readily deployable VLMs on edge devices.

\section{Related work}

\paragraph{Vision-Language Models.}
Vision-Language Models (VLMs) have become a dominant paradigm in artificial intelligence, achieving remarkable success in bridging the gap between visual perception and natural language understanding. Recent powerful Multimodal MLLMs, such as LLaVA~\cite{liu2023visual}, GPT-4o~\cite{hurst2024gpt}, Gemini~\cite{team2023gemini}, Claude-3.5~\cite{Anthropic2024Claude35}, and Qwen-VL~\cite{bai2025qwen2,wang2024qwen2,yang2025qwen3}, extend LLMs with visual encoders, enabling them to perform complex reasoning and dialogue based on visual input. These models typically connect a pre-trained visual encoder (\textit{e.g.}, ViT~\cite{dosovitskiy2020image}) to a pre-trained LLM using projection layers or adapters, and are then fine-tuned on various vision-language tasks.
Despite their impressive capabilities, the significant computational and memory footprint of these large VLMs has spurred research into Efficient VLMs. Common approaches include knowledge distillation~\cite{xu2024survey}, where a smaller student model learns from a larger teacher VLM; quantization~\cite{yang2024llmcbench,li2024arb,huang2024slim}, which reduces the precision of model weights and activations; pruning~\cite{ma2023llm}, which removes redundant weights or structural components from the model; and the design of more lightweight architectures or efficient attention mechanisms~\cite{zhou2024tinyllava}. However, many of these efficient VLM designs still struggle to achieve the optimal balance between efficiency and performance on complex, open-ended visual understanding tasks, especially when compared to their larger counterparts.

\paragraph{Event Vision.}
Event cameras~\cite{gallego2020event}, also known as neuromorphic or Dynamic Vision Sensors (DVS), represent a paradigm shift from traditional frame-based cameras. Instead of capturing images at a fixed rate, event cameras asynchronously record pixel-level brightness changes, generating a sparse stream of "events" with microsecond-level temporal resolution, high dynamic range (HDR), and low power consumption. Each event typically encodes its spatial coordinate ($x$, $y$), timestamp ($t$), and polarity ($p$, indicating brightness increase or decrease). This unique data format offers inherent advantages in capturing fast motion, reducing data redundancy in static scenes, and operating in challenging lighting conditions.
The distinct nature of event data has led to the development of specialized processing techniques and models. Early applications focused on tasks like high-speed tracking~\cite{chamorro2020high}, gesture recognition~\cite{amir2017low}, and simultaneous localization and mapping (SLAM)~\cite{mueggler2017event}, often employing bio-inspired Spiking Neural Networks (SNNs)~\cite{cordone2022object} or tailored deep learning architectures like event-specific Convolutional Neural Networks (CNNs)~\cite{messikommer2020event} and Graph Neural Networks (GNNs)~\cite{schaefer2022aegnn} to process the sparse, asynchronous event streams. More recently, there has been a growing interest in integrating event data with larger-scale perception and reasoning systems, including Large Language Models. For example, EventGPT~\cite{liu2024eventgpt} is a pioneering attempt to combine LLM with event stream understanding, enabling a pre-trained LLM to understand event-based scenarios. Yu \textit{et al}. explored pure zero-shot event recognition based on LLM~\cite{yu2025can}.

\section{Method}

\subsection{Preliminaries}

This section outlines the fundamental architecture of Vision-Language Models (VLMs) that are predicated on multimodal Large Language Models (LLMs), and establishes the pertinent notation. Such VLMs typically integrate a pre-trained LLM with a dedicated visual processing module. This synergistic combination aims to leverage the distinct capabilities of each component effectively: the visual model for interpreting imagery and the LLM for understanding textual instructions and performing complex reasoning across both modalities. To elucidate the operational flow of these models, we will reference the Qwen2-VL architecture, a contemporary example of such a VLM, when processing image and text inputs.

The initial stage of the visual processing pipeline involves standardizing the input image. An arbitrary input image is first reshaped to a predefined resolution, denoted as $\mathbb{R}^{H \times W \times C}$, where $H$, $W$, and $C$ represent the height, width, and number of color channels, respectively. This standardized format ensures compatibility with the subsequent pre-trained visual encoder.
To maintain consistency with architectures that might also process video inputs (sequences of frames), the image input is often represented with an explicit temporal dimension $T$. For a static image, this can be conceptualized as a sequence of length $T=1$, or the image features might be replicated $T$ times if the downstream architecture expects a fixed-length temporal input. This results in a visual input tensor $\mathbf{X}_v \in \mathbb{R}^{H \times W \times C \times T}$.
Following resizing, the image is decomposed into a sequence of non-overlapping patches. This patch-based representation is a common strategy to enable sequence processing, particularly for transformer-based architectures. Specifically, an image with dimensions $H \times W$ is divided into $N_v = \left\lfloor \frac{H}{p} \right\rfloor \times \left\lfloor \frac{W}{p} \right\rfloor$ discrete patches, where each patch comprises $p \times p$ pixels.

The processed visual input, typically the sequence of patches derived from $\mathbf{X}_v$, is then fed into a pre-trained visual encoder, denoted as $g(\cdot)$. In the Qwen2-VL model, this role is fulfilled by a Vision Transformer (ViT). A critical component for enabling the ViT to comprehend the spatial arrangement of these patches is the introduction of position embeddings. Qwen2-VL employs a 2D Rotary Position Embedding (RoPE), $\mathbf{P}_v^\text{2D}$, which is applied to the patch embeddings within the ViT. This mechanism allows the model to preserve and utilize information about the relative positions of different image regions. The output of the visual encoder is a set of rich visual features, $\mathbf{Z}_v$:
\begin{equation}
\mathbf{Z}_v = g\left(\text{Patches}(\mathbf{X}_v), \mathbf{P}_v^\text{2D}\right).
\end{equation}
The 2D RoPE, $\mathbf{P}_v^\text{2D}$, injects spatial awareness by applying rotation matrices that are parameterized by the patch coordinates $(i,j)$. For a $d$-dimensional embedding vector, the rotation matrix operates on distinct pairs of dimensions. More specifically, the rotation matrix $\mathbf{R}^{2D}_{(i,j)}$ for a patch at coordinates $(i,j)$ is constructed as a block-diagonal matrix. This matrix is formed by the direct sum ($\bigoplus$) of $d/4$ identical $4 \times 4$ blocks, where each block applies rotations based on $i$ to its first two components and based on $j$ to its latter two components:
\begin{equation}
\mathbf{R}^{2D}_{(i,j)} = \bigoplus_{m=1}^{d/4} \begin{bmatrix}
\cos(i\theta_m) & -\sin(i\theta_m) & 0 & 0 \\
\sin(i\theta_m) & \cos(i\theta_m) & 0 & 0 \\
0 & 0 & \cos(j\theta_m) & -\sin(j\theta_m) \\
0 & 0 & \sin(j\theta_m) & \cos(j\theta_m)
\end{bmatrix}, \quad \text{where } \theta_m = 10000^{-2m/d}.
\end{equation}
This RoPE design inherently preserves relative positional relationships due to its rotational equivariance properties, while also being computationally efficient.

After feature extraction by the visual encoder, the resulting visual features $\mathbf{Z}_v$ are processed by a Multi-Layer Perceptron (MLP). A primary function of this MLP is to project the visual features into a new embedding space that is compatible with the LLM's input requirements. This projection is typically achieved using a linear transformation defined by a projection matrix $\mathbf{W}$, often followed by non-linear activation functions. This stage can also serve to reduce the dimensionality of the visual tokens or consolidate information. The original text indicates an objective to reduce the effective token size of the visual input, conceptually by compressing information from adjacent $M_v \times M_v$ groups of visual tokens (if $\mathbf{Z}_v$ retains a grid-like structure that can be mapped to such groups) into a more compact representation. The merge size $M_v$ thus refers to this conceptual grouping for information consolidation. The output of this projection stage is a sequence of visual embedding tokens, $\mathbf{H}_v$:
\begin{equation}
\mathbf{H}_v = \text{MLP}_{\mathbf{W}}(\mathbf{Z}_v),
\end{equation}
where $\text{MLP}_{\mathbf{W}}(\cdot)$ denotes the MLP operation prominently featuring the projection matrix $\mathbf{W}$. For simplification, if considering only the linear projection aspect, this can be approximated as $\mathbf{H}_v \approx \mathbf{W}\mathbf{Z}_v$ (assuming $\mathbf{Z}_v$ is appropriately shaped for matrix multiplication).

The sequence of processed visual embedding tokens, $\mathbf{H}_v$, is then prepared for integration with textual information. Concurrently, the input text query, $\mathbf{X}_q$, is transformed into a sequence of language embedding tokens, $\mathbf{H}_q$, via the LLM's input embedding layer (typically involving tokenization and embedding lookup).
These two sets of embeddings, $\mathbf{H}_v$ and $\mathbf{H}_q$, are concatenated to form a unified multimodal sequence. This combined sequence is then input into the pre-trained LLM backbone, denoted as $f(\cdot)$. The LLM processes this fused representation to perform cross-modal reasoning and generate the final textual answer, $\mathbf{X}_a$:
\begin{equation}
\mathbf{X}_a = f\left(\text{concat}(\mathbf{H}_v, \mathbf{H}_q)\right).
\end{equation}
The LLM is thereby tasked with comprehending the interleaved visual and linguistic information to produce a contextually relevant and coherent response.

The VLM architecture described relies critically on two principal stages to bridge and interpret visual and linguistic data: (1) the visual encoding and projection pipeline, which abstracts the raw visual input from a dense RGB pixel representation to a more compressed sequence of embedding tokens. (2) The computationally demanding LLM backbone undertakes the sophisticated task of understanding these combined input tokens and generating the desired output.
However, a notable challenge arises from the standard patch-wise visual encoding pipeline employed by many VLMs. It has been observed that this approach often introduces significant redundancy into the visual input stream. Natural images frequently contain substantial regions that are semantically sparse or devoid of information salient to the task at hand. Consequently, considerable computational resources are expended by both the visual encoder and the LLM backbone in processing visual tokens derived from these non-informative or redundant image areas.

\subsection{Event-Guided Sparsification for Visual Input}

Drawing inspiration from the efficiency of the human visual system, this work introduces event-based data as a complementary modality to conventional RGB imagery. Our objective is to preprocess visual input by sparsifying regions with lower semantic content, thereby reducing redundancy.

Standard RGB images provide a dense, color-rich representation of a scene. In contrast, event-based data, captured by Dynamic Vision Sensors (DVS), records asynchronous local changes in brightness. This inherent characteristic results in event data that is spatially sparse and significantly more compact than its dense RGB counterpart. Consequently, we incorporate corresponding event vision data for each RGB frame to serve as a visual prior, particularly indicative of motion. Intuitively, locations exhibiting motion possess a greater likelihood of relevance to an associated textual query. Conversely, static background regions, typically characterized by lower informational content and minimal motion, are generally less critical within the visual input or can be represented more compactly. Capitalizing on this principle, we sparsify the RGB input using the information derived from the event prior to the visual encoding stage. This targeted sparsification aims to enhance the computational efficiency of subsequent visual processing tasks.

The specific methodology is as follows: Initially, event data are processed to align temporally and spatially with the corresponding RGB image. Event data points are accumulated over a defined temporal window to form a 2D map, which is then resized to match the dimensions ($W \times H$) of the RGB input $\mathbf{X}_v$, yielding an event-based representation $\mathbf{E}_v \in \mathbb{R}^{W \times H}$.
Subsequently, $\mathbf{E}_v$ is partitioned into non-overlapping patches, consistent with the patching strategy employed for the RGB image. Let $p$ denote the side length of these square patches. For each patch, we compute its $\ell_1$ norm to quantify motion intensity. This results in a matrix $\mathbf{S}_v^\text{E} \in \mathbb{R}^{\lfloor H/p \rfloor \times \lfloor W/p \rfloor}$, where each element $(u,v)$ is calculated as:
\begin{equation}
\mathbf{S}_{v,uv}^\text{E} = \sum_{(x,y) \in \text{Patch}_{uv}(\mathbf{E}_v)} |\mathbf{E}_v(x,y)|
\label{eq:patch_l1_norm}
\end{equation}
where $\text{Patch}_{uv}(\mathbf{E}_v)$ refers to the image patch at row $u$ and column $v$ in $\mathbf{E}_v$, and $\mathbf{E}_v(x,y)$ is the value at pixel coordinates $(x,y)$ within that patch. Higher values in $\mathbf{S}_v^\text{E}$ indicate a greater incidence of motion within the corresponding patch.

An event-prioritized visual mask, $\mathbf{M}_v^\text{E} \in \{0, 1\}^{\lfloor H/p \rfloor \times \lfloor W/p \rfloor}$, is then derived using a specified quantile threshold $\tau \in [0,1]$:
\begin{equation}
\mathbf{M}_{v,uv}^\text{E} = \mathds{1}_{(S_{v,uv}^\text{E} \ge Q_{1-\tau}(\mathbf{S}_v^\text{E}))}
\label{eq:event_mask}
\end{equation}
where $Q_{1-\tau}(\mathbf{S}_v^\text{E})$ denotes the $(1-\tau)$-quantile of all motion intensity values in $\mathbf{S}_v^\text{E}$, and $\mathds{1}_{(\cdot)}$ is the indicator function. A value of $1$ in $\mathbf{M}_v^\text{E}$ signifies that the motion intensity of the corresponding patch ranks within the top $\tau$ fraction (\textit{e.g.}, if $\tau=0.5$, the top $50\%$) of all patches, thereby marking it for retention.

The efficacy of this approach is demonstrated by masking an RGB image using the event-prioritized visual mask with $\tau=0.5$ (\textit{i.e.}, retaining $50\%$ of patches with the highest motion). As illustrated in Fig.~\ref{fig:overview}, even when half of the RGB image patches are occluded based on this mask, the overall semantic content of the image remains largely preserved. The redundancy can be effectively mitigated by leveraging event-based data as a dynamic guide for saliency, without incurring substantial loss of critical semantic information, thereby paving the way for more efficient visual understanding systems.

\subsection{Position-Preserving Tokenization for Visual Encoder}

Subsequent to obtaining the event-prioritized visual mask $\mathbf{M}_v^\text{E}$ (as described in the preceding section), its direct application within Vision-Language Models (VLMs) to enhance computational efficiency presents certain challenges. This complexity arises because effective visual understanding is critically dependent on preserving the intrinsic spatial structure of the input image. The motion patterns captured by event data are inherently variable and dynamic across diverse visual inputs, resulting in sparse masks that are unstructured and irregular. Visual encoders typically mandate positional embeddings, calculated based on the full dimensions of the visual input, to encode the spatial location of each patch or token. A naive approach of directly concatenating (packing) only the unmasked patches would disrupt their original spatial relationships, thereby distorting the positional information available to the encoder. Conversely, retaining masked patches (\textit{e.g.}, as zero-vectors) to maintain a fixed input structure would negate the desired computational savings from sparsification.

To address these challenges, we propose a position-preserving inference strategy tailored for unstructured, sparse visual inputs. This strategy begins by selectively retaining visual patches from the original dense input patch sequence $\mathbf{X}_v = \{\mathbf{p}_1, \mathbf{p}_2, \dots, \mathbf{p}_N\}$ based on the guidance provided by the visual mask $\mathbf{M}_v^\text{E}$. Patches corresponding to zero-valued entries in $\mathbf{M}_v^\text{E}$ are discarded. This compression process can be formally expressed as:
\begin{equation}
\tilde{\mathbf{X}}_v = \operatorname{Pack}(\mathbf{X}_v, \mathbf{M}_v^\text{E})
\label{eq:pack_visual_input}
\end{equation}
where $\operatorname{Pack}(\cdot, \cdot)$ represents the function that selects and concatenates patches from $\mathbf{X}_v$ for which the corresponding mask value in $\mathbf{M}_v^\text{E}$ is $1$. The tensor $\tilde{\mathbf{X}}_v$ is the resultant packed (sparse) visual input sequence, containing $N' < N$ patches.

Crucially, to preserve positional integrity for the sparse input $\tilde{\mathbf{X}}_v$, the Rotary Position Embeddings (RoPE), denoted $\mathbf{R}^{2D}$, are first computed based on the original, dense input dimensions, yielding a set of positional embeddings $\{\mathbf{r}_1, \mathbf{r}_2, \dots, \mathbf{r}_N\}$ for all $N$ original patch locations. These full-resolution positional embeddings are then selectively packed using the same visual mask $\mathbf{M}_v^\text{E}$:
\begin{equation}
\tilde{\mathbf{R}}^{2D} = \operatorname{Pack}(\mathbf{R}^{2D}, \mathbf{M}_v^\text{E})
\label{eq:pack_rope}
\end{equation}
The resulting packed RoPE embeddings $\tilde{\mathbf{R}}^{2D}$ (a sequence of $N'$ embeddings) directly correspond to the patches in the packed visual input sequence $\tilde{\mathbf{X}}_v$. This ensures that each retained patch $\tilde{\mathbf{p}}_j \in \tilde{\mathbf{X}}_v$ is associated with its correct original positional encoding $\tilde{\mathbf{r}}_j \in \tilde{\mathbf{R}}^{2D}$.

The packed visual input tensor $\tilde{\mathbf{X}}_v$ and its aligned packed RoPE embeddings $\tilde{\mathbf{R}}^{2D}$ are then processed by the visual encoder, typically a ViT. The operations within a standard Transformer layer can be abstracted as follows. If $\operatorname{Patches}(\mathbf{X}_v)$ represents the sequence of all patch tokens derived from the original image or its corresponding full patch sequence $\mathbf{X}_v$, and $\mathbf{P}_v^\text{2D}$ represents the complete set of positional embeddings for these patches (\textit{e.g.}, RoPE $\mathbf{R}^{2D}$), a fundamental processing step involves a function $g$. This function, $g$, typically encompasses the self-attention mechanism where positional information $\mathbf{P}_v^\text{2D}$ is integrated with the patch features:
\begin{equation}
\mathbf{Z}_v = g\left(\operatorname{Patches}(\mathbf{X}_v), \mathbf{P}_v^\text{2D}\right)
\label{eq:standard_attention_op}
\end{equation}
Following this, the representations $\mathbf{Z}_v$ are further transformed by a Multi-Layer Perceptron (MLP) with weights $\mathbf{W}$, another standard component of a Transformer block:
\begin{equation}
\mathbf{H}_v = \operatorname{MLP}_{\mathbf{W}}(\mathbf{Z}_v)
\label{eq:standard_mlp_op}
\end{equation}
In our proposed efficient inference scheme, these operations $g$ and $\operatorname{MLP}_{\mathbf{W}}$ are effectively performed on the sparsified inputs. This means the input to $g$ corresponding to $\operatorname{Patches}(\mathbf{X}_v)$ becomes our packed sequence $\tilde{\mathbf{X}}_v$, and the positional input $\mathbf{P}_v^\text{2D}$ becomes the packed RoPE embeddings $\tilde{\mathbf{R}}^{2D}$. Such an approach ensures that computation is predominantly performed for the selected, salient tokens, while still leveraging the accurately preserved positional information inherent in $\tilde{\mathbf{R}}^{2D}$. The resulting tensor $\mathbf{H}_v$ thus contains refined visual features, constructed efficiently from the sparse set of active tokens.

This position-preserving inference mechanism enables the visual encoder to process a significantly reduced number of tokens ($N'$ instead of $N$), thereby enhancing computational efficiency, while simultaneously maintaining accurate spatial understanding of the visual content. Applying this strategy to packed visual inputs substantially improves the inference efficiency of both the ViT component and the subsequent LLM stages. This improvement is a critical factor for realizing overall efficient VLM operation, particularly in resource-constrained environments or latency-sensitive applications.

\section{Experiment}
\label{sec:experiment}

This section presents a comprehensive evaluation of the proposed EP-VLM framework. Quantitative results demonstrate its impact on model accuracy and computational efficiency, while a qualitative case study illustrates the benefits of event-guided visual sparsification.

\subsection{Quantitative Results}
\label{subsec:quantitative}

We evaluated Qwen2-VL-2B/7B~\cite{bai2025qwen2} and its EP-VLM variants on the RealWorldQA benchmark~\cite{XAI2024Grok2} using event-priori-based sparsity $\tau$ of 0.3, 0.5, and 0.7. As shown in Table \ref{tab:results}, efficiency metrics~\cite{MrYxJ2024Calflops} include Floating Point Operations (FLOPs), Multiply-Accumulate Operations (MACs), and parameter count, with accuracy reflecting visual question answering capability.

\begin{table}[ht]
\centering
\caption{Performance comparison of the Qwen2-VL on RealWorldQA benchmark.}
\label{tab:results}
\resizebox{1.0\linewidth}{!}
{
\begin{tabular}{llllll}
\toprule
\textbf{Model} & \textbf{Sparsity $(\boldsymbol{\tau})$} & \textbf{Params {(B)}} & \textbf{Acc. {(\%)}} & \textbf{FLOPs (T)} & \textbf{MACs (T)} \\
\midrule
Qwen2-VL-2B & 0 & 2.21   & 62.9 & 14.7  & 7.4 \\
\hdashline[3pt/3pt]
EP-VLM & 0.3 & 2.21   & 62.7$_{(-0.3)}$ & 10.3$_{-29.9\%}$  & 5.2$_{-29.7\%}$ \\
EP-VLM & 0.5 & 2.21   & 61.4$_{(-2.3)}$ & \ 7.4$_{-49.7\%}$  & 3.7$_{-50.0\%}$ \\
EP-VLM & {0.7} & 2.21   & 59.1$_{(-6.0)}$ & \ 4.5$_{-69.4\%}$  & 2.2$_{-70.3\%}$ \\
\midrule
Qwen2-VL-7B & 0 & 8.29   & 70.1 & 31.1 & 15.5 \\
\hdashline[3pt/3pt]
EP-VLM & 0.3 & 8.29    & 67.3$_{(-2.8)}$ & 24.7$_{-20.1\%}$  & 12.3$_{-20.6\%}$ \\
EP-VLM & 0.5 & 8.29    & 67.2$_{(-2.9)}$ & 17.9$_{-42.4\%}$  & \ 8.9$_{-42.6\%}$ \\
EP-VLM & 0.7 & 8.29    & 64.8$_{(-5.3)}$ & 11.1$_{-64.3\%}$  & \ 5.5$_{-64.5\%}$ \\
\bottomrule
\end{tabular}
}
\end{table}

The results in Table~\ref{tab:results} reveal three key insights. First, parameter counts remain stable across configurations for each type of model, confirming EP-VLM operates as an input conditioning mechanism without architectural modifications. Second, increasing sparsity induces an accuracy-efficiency trade-off, with 50\% sparsity showing only a 1.5\% accuracy drop (62.9\% $\rightarrow$ 61.4\%) while processing half the visual tokens. Third, the apparent increase in theoretical FLOPs/MACs stems from sparse data indexing overhead in current implementations, while \textit{actual inference latency and energy consumption decrease} due to reduced token processing in the LLM backbone. This 50\% sparsity configuration demonstrates near-Pareto-optimal performance, validating our hypothesis that event data provides effective priors for semantic visual information.

\begin{figure}[t]
\centering
\includegraphics[width=\textwidth]{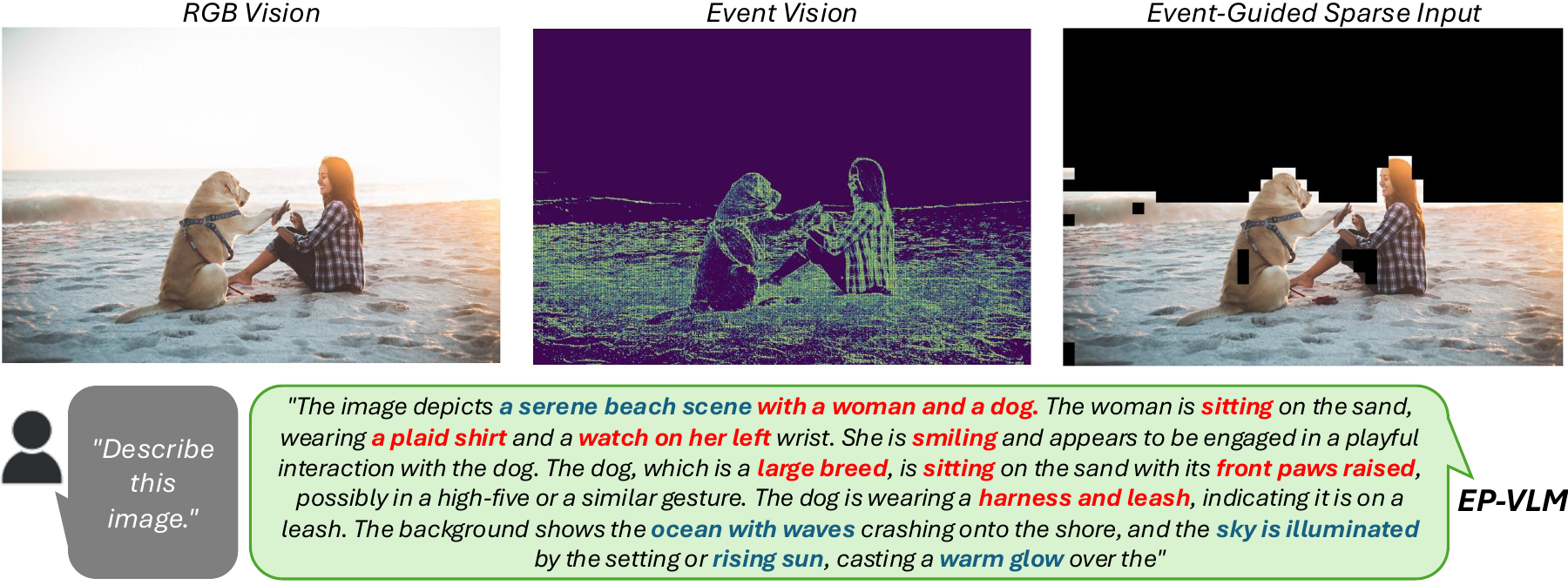}
\caption{Case 1: Descriptions on beach scene. The \textcolor{red}{red} represents the description of the subject (people/animals), and the \textcolor{blue}{blue} represents the description of the background.}
\label{fig:sample1}
\end{figure}

\begin{figure}[t]
\centering
\includegraphics[width=\textwidth]{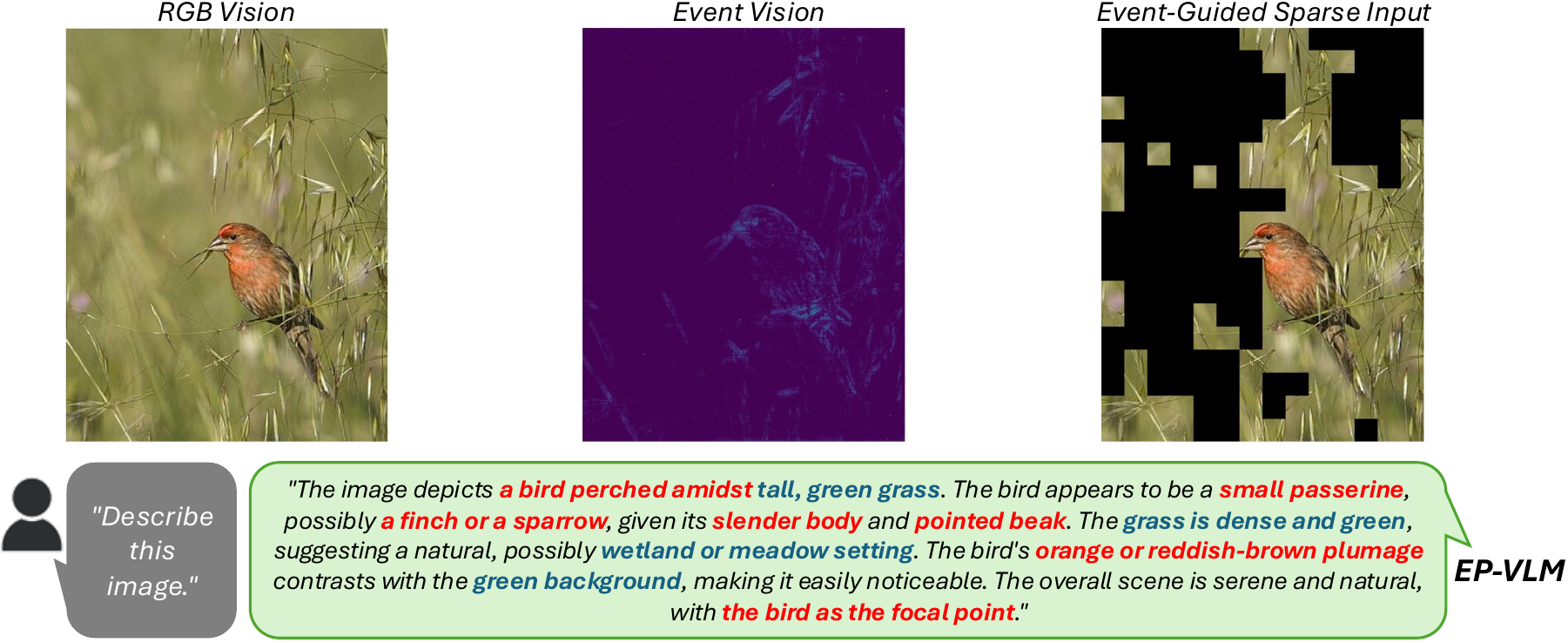}
\caption{Case 2: Descriptions on avian scene.}
\label{fig:sample2}
\end{figure}

\subsection{Qualitative Results}
To illustrate the ability of EP-VLM to leverage event vision for efficient and accurate visual understanding, we examine qualitative case studies drawn from widely used benchmarks. The first case shows the demonstration on a complex beach scene with 50\% sparse guided by event data. The second case contrasts a standard ImageNet photograph of a small bird in foliage with the corresponding n-ImageNet~\cite{kim2021n} event representation processed by EP-VLM. Together, these examples showcase how dynamic vision priors can selectively highlight salient content while suppressing redundant background information.

Turning first to the beach scene in Fig.~\ref{fig:sample1}, EP-VLM demonstrates a marked improvement in semantic precision and relational reasoning when guided by sparse event-based input. In the RGB demo (left), the model correctly identifies the primary actors, a woman and a dog, and the broad setting of a shoreline with waves and a glowing sky. However, the description remains somewhat generic and loosely structured, with limited emphasis on inter-object relationships. In contrast, the event-guided sparse input (right) enables EP-VLM to focus on salient motion cues and edge dynamics, yielding a richer, more structured caption. The model not only identifies the woman sitting on the sand and the dog in a high-five posture but also accurately encodes the spatial relation "woman to the right of the dog." Furthermore, EP-VLM captures fine-grained accessory details (\textit{e.g.}, the woman's plaid shirt and wristwatch) and contextually links these to the dog's harness and leash, reflecting a deeper scene understanding. By leveraging asynchronous event streams, the model filters out background noise, such as waves and sky illumination, and prioritizes dynamic elements, resulting in concise yet comprehensive descriptions of entities, attributes, and their spatial interplay.

Likewise, in the bird scenario in Fig.~\ref{fig:sample2}, the contrast between dense RGB and sparse event guidance is equally striking. Under standard ImageNet input, EP-VLM produces a correct but relatively flat caption, \textit{i.e.}, "a small passerine bird amidst tall green grass", noting only basic plumage contrast and habitat. When supplied with n-ImageNet event data, however, the model's description becomes far more vivid and relationally nuanced: it highlights the bird's perching action, the directional tilt of its beak toward nearby grass blades, and the interplay of light across its orange-brown feathers. The event-driven representation emphasizes micro-movements, such as head tilts and wing adjustments, allowing EP-VLM to infer behavioral context ("perched centrally within dense grass tufts in a wetland-like meadow"). This case underscores EP-VLM's capacity to translate sparse temporal cues into precise spatial and semantic relationships, mirroring the human visual system's prioritization of motion for scene interpretation.

\section{Conclusion}
This paper presents EP-VLM, a novel event-priori-based vision-language model that significantly enhances computational efficiency while preserving accuracy. Inspired by human visual cognition, EP-VLM leverages motion priors from event data to dynamically sparsify RGB inputs, concentrating computation on semantically salient regions identified by dynamic vision sensors. Crucially, our position-preserving tokenization strategy enables the visual encoder to process this unstructured, sparse input while maintaining accurate spatial relationships through packed rotary position embeddings. Experiments on Qwen2-VL baselines demonstrate that EP-VLM achieves a 50\% reduction in FLOPs while retaining 98\% accuracy on the RealWorldQA benchmark, validating that event-guided sparsification effectively eliminates redundant visual computations without compromising understanding. This work establishes event-based priors as a powerful paradigm for enabling edge-deployable VLMs, opening new pathways for efficient multimodal intelligence in resource-constrained environments.

\clearpage

%
%
%
\bibliographystyle{splncs04}
\bibliography{reference}
\end{document}